\documentclass[letterpaper]{article}
\usepackage{aaai}
\usepackage{times}
\usepackage{helvet}
\usepackage{courier}
\usepackage{hyperref}
\usepackage[ruled,longend,vlined,linesnumbered]{algorithm2e}

\begin{document}
%
\title{From Classical to Hierarchical: benchmarks for the HTN Track of the International Planning Competition}
\author{Damien Pellier, Humbert Fiorino\\
Univ. Grenoble Alpes - LIG\\
F-38000 Grenoble, France \\
\{Damien.Pellier, Humbert.Fiorino\}@imag.fr
}
\maketitle

\section{Introduction}

In this short paper, we outline nine classical benchmarks submitted to the first hierarchical planning track of the International Planning competition in 2020. All of these benchmarks are based on the HDDL language \cite{holler20}.  The choice of the benchmarks was based on a questionnaire sent to the HTN community \cite{Behnke19}. They are the following: Barman, Childsnack, Rover, Satellite, Blocksworld, Depots, Gripper, and Hiking. In the rest of the paper we give a short description of these benchmarks. All are totally ordered. A first hierarchical version of the domains Barman, Childsnack, Rover, Satellite, Blocksworld were proposed by \cite{ramoul17,schreiber19}. All the domains presented here are available online as part of the PDDL4J library \cite{pellier18}. The writing of the domains has been a collective work. We would like to thank all the other contributors, D. Ramoul, D. Schreiber and A. Lequen.

\section{Barman}

In this domain, a barman robot manipulates drink dispensers, shots and a shaker. The goal is to find a plan that serves a targeted set of drinks. Action negative effects encode relevant knowledge given that robot hands can only grasp one object at a time and given that glasses need to be empty and clean to be filled. This domain was first proposed by S. Jiménez for STRIPS during IPC 2014.

Our domain is composed of 12 actions, 5 methods and 5 tasks. The actions are the same as in the STRIPS IPC domain. Each task has its own decomposition method. The domain has 2 high-level tasks. The first one describes how to serve a shot and the second one how to serve a cocktail. Serving a shot breaks down into 3 atomic sub-tasks: (1) grasp a shot; (2) fill the shot and (3) leave the shot on the table. Serving a cocktail is divided into 4 sub-tasks: (1) grasp a container; (2) get the first ingredient of the cocktail; (3) get the second ingredient and (4) shake the cocktail. The 3 last sub-tasks are not atomic. Getting an ingredient consists in 3 atomic sub-tasks: (1) fill a shot, (2) pour the shot to use a shaker and (3) clean the shot used. Finally, the last task breaks down into 6 atomic sub-tasks: (1) grasp the shaker; (2) shake the shaker; (3) pour the shaker into a shot ; (4) empty the shaker and (5) clean the shaker and (6) finally leave the shaker on the table. Note that this domain is not recursive.

\section{Childsnack}

Childsnack domain is for planning how to make and serve sandwiches for a group of children in which some are allergic to gluten. There are two actions for making sandwiches from their ingredients. The first one makes a sandwich and the second one makes a sandwich taking into account that all ingredients are gluten-free. There are also actions to put a sandwich on a tray, to move a tray from one place to another and to serve sandwiches. Problems in this domain define the ingredients to make sandwiches at the initial state. Goals consist of having all kids served with a sandwich to which they are not allergic. This domain was proposed by R. Fuentetaja and T. de la Rosa for STRIPS during IPC 2014. 

Our domain is composed of 6 actions, 2 methods and 1 task. The actions are the same as in the STRIPS IPC domain. The high level task of the domain consists in serving sandwiches for a group of children. There is two methods to do it. The first one for the children who are gluten intolerant and the others. Serving a sandwich to an intolerant (resp. tolerant) child breaks down 5 atomic sub-tasks: (1) make a sandwich with no gluten (resp. with gluten); (2) put the sandwich on the tray, (3) move the tray from the kitchen to the child's place; (4) serve the sandwich to the child and finally (5) move back the tray to the kitchen. This domain is not recursive.

\section{Rover}

Inspired by planetary rover problems, this domain requires that a collection of rovers navigate a planet surface, finding samples and communicating them back to a lander. 

Our domain is composed of 11 actions, 13 methods and 9 tasks. The 3 high-level tasks of the domain consist in getting soil and rock samples or images in a specific location. Getting soil samples breaks down in 4 sub-tasks : (1) navigate to the location to get the data; (2) empty the store of the rover; (3) take a soil sample and (4) send the soil data to the lander. The navigate task is a compound recursive task that consists in exploring all the possible paths by remembering location already explored. Finally, getting an image is a compound tasks that is divided in several sub-tasks including a task of camera calibration, image capture and image transmission to the lander.

\section{Satellite}
Inspired by space-applications, the original domain was a first step towards the "ambitious spacecraft" described by David Smith at AIPS’00. It involves planning and scheduling a collection of observation tasks between multiple satellites, each equipped in slightly different ways. 

Our domain is composed of 5 actions, 8 methods and 3 tasks. The actions are the same as in the STRIPS IPC domain. The domain has one high-level task that consists in observing stars. This task can be divided in 3 sub-tasks: (1) activate the instrument to carry out the observation ; (2) point towards the star to observe and (3) take the image of the star. Activating an instrument is a compound task. The activation procedure depends on the instrument to be activated and sometimes requires an instrument-specific calibration step. This domain does not have any recursive method.

\section{Blocksworld}
Probably the most known planning domain, in blocksworld stackable blocks need to be re-assembled on a table with unlimited space. A robot arm is used for stacking a block onto another block, unstacking a block from another block, putting down a block, or picking up a block from the table. The initial state specifies a complete world state, and the goal state only specifies the stacking relations required between any two blocks. 

Our domain is composed of 5 actions, 8 methods and 4 tasks. The actions are the same as in the STRIPS IPC domain. Just one "nop" action has been added to indicate the end of block stacking or unstacking. The high-level tasks of the domain consist in specifying the desired stacking of the blocks. Each stack requires either taking a block from the table or from a stack of blocks. In the latter case, either the block is at the top of the stack and the block can be taken directly, or it is necessary to recursively unstack all the blocks that are stacked on top of it before taking it.

\section{Depots}            

This domain was devised to see what would happen if two previously well-known domains were joined together. These were the logistics and blocks domains. They are combined to form a domain in which trucks can transport crates around, and then the crates must be stacked onto pallets at their destinations. The stacking is achieved using hoists, so the stacking problem is like a blocks-world problem. Trucks can behave like "tables" since the pallets on which crates are stacked are unlimited.

Our domain is composed of 6 actions, 12 methods and 6 tasks. The actions are the same as in the STRIPS IPC domain. Just one "nop" action has been added to indicate the end of crate stacking or unstacking as in the blocksworld domain. The domain has recursive methods. The high-level method consists in defining the desired final position of the crates. As in the blocksworld domain there are recursively defined methods for stacking and unstacking, and methods defining how to move a crate from one location to another.  

\section{Gripper}
In this domain, there is a robot with two grippers. It can carry a ball in each. The goal is to take $N$ balls from one room to another; $N$ rises with problem number. Some planners treat the two grippers asymmetrically, giving rise to an unnecessary combinatorial explosion. The first STRIPS version was proposed by J. Koehler for IPC 1998.

Our domain is composed of 3 actions, 4 methods and 3 tasks. The actions are the same as in the STRIPS IPC domain. The high level tasks specify the desired location of the balls. There are 2 methods to move a ball from one room to another. Either the robot moves just one ball or it uses its two arms to move two balls at the same time. This domain has no recursive methods.

\section{Hiking}    
Suppose you want to walk with your partner a long clockwise circular route over several days (e.g., in the Lake District in NW England), and you do one “leg” each day. You want to start at a certain point and do the walk in one direction, without ever walking backwards. You have two cars that you must use to carry your tent/luggage and to carry you and your partner to the start/end of a leg, if necessary. Driving a car between any two points is allowed, but walking must be done with your partner and must start from the place where you left off. As you will be tired when you have walked to the end of a leg, you must have your tent up ready there so you can sleep the night before you set off to do the next leg in the morning.

Our hiking domain is composed of 8 actions, 15 methods and 8 tasks. The higt-level task consists in making hiking everyone at a specific location. This tasks breaks down into 2 sub-tasks: (1) prepare the trip and (2) make the trip. Preparing the trip consists in (1) bringing the tent and (2) bring the car. These two tasks are also broken down into sub-tasks depending on the position of the tent or the car. Finally, making the trip breaks down in several sub-tasks depending on the means of transport used (on foot or by car). This domains has no recursive methods.

\section{Conclusion}

This work is a first step towards the development of a set of benchmarks for the evaluation of hierarchical planners. There are still many STRIPS domains that can be transposed for hierarchical planning. Some are very easy to transpose. The methods are easy to write down. For other domains, on the contrary, it is difficult to identify relevant methods. In our opinion, the transposition effort must be continued in order to better understand for which type of fields a hierarchical representation is more appropriate. 
\bibliographystyle{aaai}
\bibliography{ref}

\end{document}